\documentclass[acmtog, authorversion, nonacm]{acmart}

\AtBeginDocument{%
  \providecommand\BibTeX{{%
    \normalfont B\kern-0.5em{\scshape i\kern-0.25em b}\kern-0.8em\TeX}}}

\acmJournal{FACMP}

\usepackage{xfp}
\usepackage{wrapfig}
\usepackage[percent]{overpic}
\usepackage{mathtools}
\usepackage{booktabs}
\usepackage{amsmath}
\usepackage{relsize}
\usepackage{xcolor}
\usepackage{caption}
\usepackage{dsfont}
\usepackage{enumitem}
\usepackage{subcaption}
\usepackage{tikz}
\usetikzlibrary{tikzmark,calc}
\usepackage{pifont}
\usepackage{stackengine}
\usepackage{bbding}
\usepackage[ruled,vlined]{algorithm2e}

\newif\ifdraft
\draftfalse

\ifdraft
\newcommand{\okc}[1]{{\color{red}[\textbf{OK:} #1]}}
\newcommand{\vpc}[1]{{\color{teal}[\textbf{VP:} #1]}}
\newcommand{\dcc}[1]{{\color{orange}[\textbf{DC:} #1]}}
\newcommand{\dlc}[1]{{\color{blue}[\textbf{DL} #1]}}

\else
\newcommand{\okc}[1]{}
\newcommand{\vpc}[1]{}
\newcommand{\dcc}[1]{}
\newcommand{\dlc}[1]{}

\fi

\definecolor{mygray}{RGB}{140, 140, 140}

\definecolor{dashedpurple}{RGB}{118,0,103}
\definecolor{dashedorange}{RGB}{181,100,13}

\DeclareMathOperator*{\argmax}{arg\,max}

\newcommand{\etal}{{\it et al.} }

\newcommand{\Wspace}{\mathcal{W}}
\newcommand{\figref}[1]{Figure~\ref{#1}}

\newcommand{\algoref}[1]{Alg.~\ref{#1}}

\setcitestyle{nosort,square}
\begin{document}

\title{Multi-level Latent Space Structuring for Generative Control}

\author{Oren Katzir}
\affiliation{\institution{Tel Aviv University} }
\author{Vicky Perepelook}
\affiliation{\institution{Tel Aviv University} }
\author{Dani Lischinski}
\affiliation{\institution{Hebrew University of Jerusalem}} 
\author{Daniel Cohen-Or}
\affiliation{\institution{Tel Aviv University} }

\begin{abstract}
Truncation is widely used in generative models for improving the quality of the generated samples, at the expense of reducing their diversity.
We propose to leverage the StyleGAN generative architecture to devise a new truncation technique, based on a decomposition of the latent space into clusters, enabling customized truncation to be performed at multiple semantic levels.
We do so by learning to re-generate $\Wspace^L$, the extended intermediate latent space of StyleGAN, using a learnable mixture of Gaussians, while simultaneously training a classifier to identify, for each latent vector, the cluster that it belongs to.
The resulting truncation scheme is more faithful to the original untruncated samples and allows better trade-off between quality and diversity.
We compare our method to other truncation approaches for StyleGAN, both qualitatively and quantitatively. 
%Furthermore, we show how such semantic clusters enable smoother and better controlled interpolation in the latent space.
\end{abstract}

\begin{teaserfigure}
  \includegraphics[width=\textwidth]{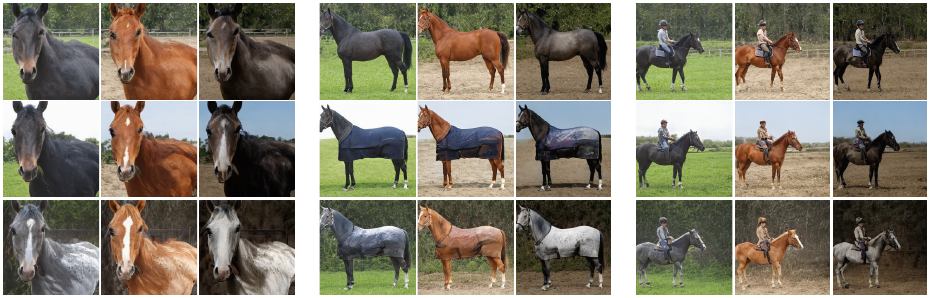}
  \caption{Multi-level structuring of StyleGAN's intermediate latent space, $\Wspace$. Specifically, we group the layers of StyleGAN's generator layers into three levels: coarse (C), medium (M) and fine (F), and learn a set of clusters for each level. Each of the three groups above corresponds to a different coarse-level cluster, where each row is from the same medium-level cluster, and each column is from the same fine-level cluster.}
  \label{fig:teaser}
\end{teaserfigure}

\maketitle

\newcommand\offsetx{.93}
\newcommand\offsety{.39}

\section{Introduction}
\label{sec:Intro}

Generative Adversarial Networks (GANs) \cite{goodfellow2014generative} have dramatically transformed unconditional image synthesis.
In particular, models based on the recent StyleGAN architecture~\cite{karras2019style,karras2020analyzing} have demonstrated an impressive ability to generate images of unprecedented photorealism.
One distinguishing feature of such generative models is that they learn a mapping from the original latent space, which is typically normally distributed, to an intermediate latent space $\mathcal{W}$, which is claimed to better mimic the distribution of the training data. 

Realizing that $\mathcal{W}$ is not a simple normal distribution suggests that it is beneficial to understand and make use of the particular structure of this space. In this paper, we explore the idea of imposing structure on the intermediate latent space in a more explicit manner. More specifically, given a pretrained generator, we learn new mappings that give rise to a collection of clusters in the latent space. Furthermore, three different collections of clusters are used to control the coarse, medium, and fine generator levels.
We show that structuring the intermediate latent space in this manner offers several benefits, particularly when dealing with image datasets that exhibit high diversity or multi-modality. 

One of the resulting advantages of discovering such a multi-level structure of the latent space is the ability to perform latent space truncation in a more sophisticated manner. Truncation essentially trades off the diversity of the generated samples for their realism. We demonstrate that performing truncation while taking into account the clustered structure of the latent space yields more realistic results in return for a significantly smaller sacrifice of diversity.

Another advantage lies in our learned mapping functions. These mapping functions allow better control on the generation process by selecting for each level of styleGAN generator a specific cluster, or multiple clusters, changing different attributes in the generated image, as shown in \figref{fig:teaser}.

We compare our method with the original styleGAN \cite{karras2019style} truncation to demonstrate the effectiveness of our approach both qualitatively and quantitatively.

\section{Related work}
\label{sec:related}
 \begin{figure*}[t]
	\centering
	\includegraphics[width=\linewidth]{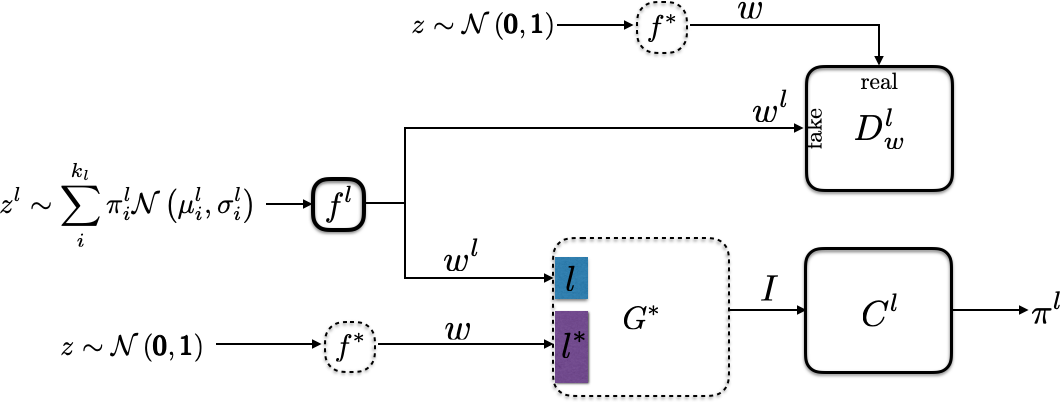}
	\caption{Single level latent architecture. We train, independently, for each level $l$ of styleGAN, a new mapping network $f^l$ and level Discriminator $D^l_w$ to generate $w^l \in \Wspace$. We encourage a multi-modal distribution by feeding the original styleGAN generator $G$, with the new learned $w^l$ only to level $l$ while other level are fed with a random $w \in \Wspace$. Simultaneously, The classifier $C^l$ is trained  to reconstruct each mode $\pi^l$. Please note that styleGAN original mapping function and image generator are pre-trained and fixed (drawn in dashed lines). }
	\label{fig:arch}
\end{figure*}

\paragraph{Multi-modal generation} features three main approaches. The common one is to construct a continuous multi-modal latent space usually in the form of a GMM, where each of the Gaussians should capture a different modality \cite{gurumurthy2017deligan, ben2018multi, pandeva2019mmgan}. 
 The second approach is to combine a continuous latent space (i.e. a Gaussian distribution) with a discrete, one-hot vector, where the latter indicates the different modalities~\cite{mukherjee2019clustergan, liu2020diverse, chen2016infogan}. The third is to over-parameterize the generative model by a mixture of generators~\cite{ghosh2018multi, khayatkhoei2018disconnected, hoang2018mgan}.
The multi modality generative learning, could be completely unsupervised, i.e. assuming the adversarial generation will tend to produce such modalities, or it is self-supervised, usually by introducing an encoder which reconstructs the selected modality index (i.e. the gaussian index, the one-hot encoding or the selected generator).
Differently, in Liu \etal~\cite{liu2020diverse} the modes in the latent space are not modeled but derived from clustering the corresponding discriminator features of the generated images.
Sendik \etal~\cite{sendik2020kmodal} extend StyleGAN to learn $k$ constants corresponding to different modes in the data.
Since the above approaches were applied on the input to the convolution layers, they usually extract very different modes and do not offer control at different semantic levels.

\paragraph{Style-based generative adversarial networks}
Style-based generative models~\cite{karras2019style, karras2020analyzing} have paved the way for a variety of applications, most of them rooted in the added control that the learned intermediate latent space $\mathcal{W}$ has on the generated model.
In contrast to $\mathcal{W}$, where the same vector is injected to all of the StyleGAN generator layers, an extended $\mathcal{W}^+$ space was found to enable embedding of a much wider range of images~\cite{abdal2019image2stylegan} into the StyleGAN latent space.
With a correct set of constraints, this allows high-quality image editing, even for out-of-training-distribution images.
Our learned modalities are also learned in an extended version of $\mathcal{W}$.

\paragraph{Evaluation of generative models}
All evaluation methods first feed-forward both real and generated images in a pre-trained network (usually inception\_v3~\cite{szegedy2016rethinking}), to extract semantically meaningful features.  The most commonly used evaluation metrics are FID \cite{heusel2017gans}, IS~\cite{salimans2016improved} and KID~\cite{arbel2018gradient}. However, since these metrics combine both quality and variety measures into a single number, it is hard to fully assess the quality of a generative model.  Therefore, the notion of precision and recall for generative models was proposed~\cite{sajjadi2018assessing}. Following the same notion, a state-of-the-art method was proposed in~\cite{kynkaanniemi2019improved}, quantifying the precision/recall by measuring how many generated/real images are within the real/generated image manifold.
\begin{figure*}
	\centering
	\includegraphics[width=\linewidth]{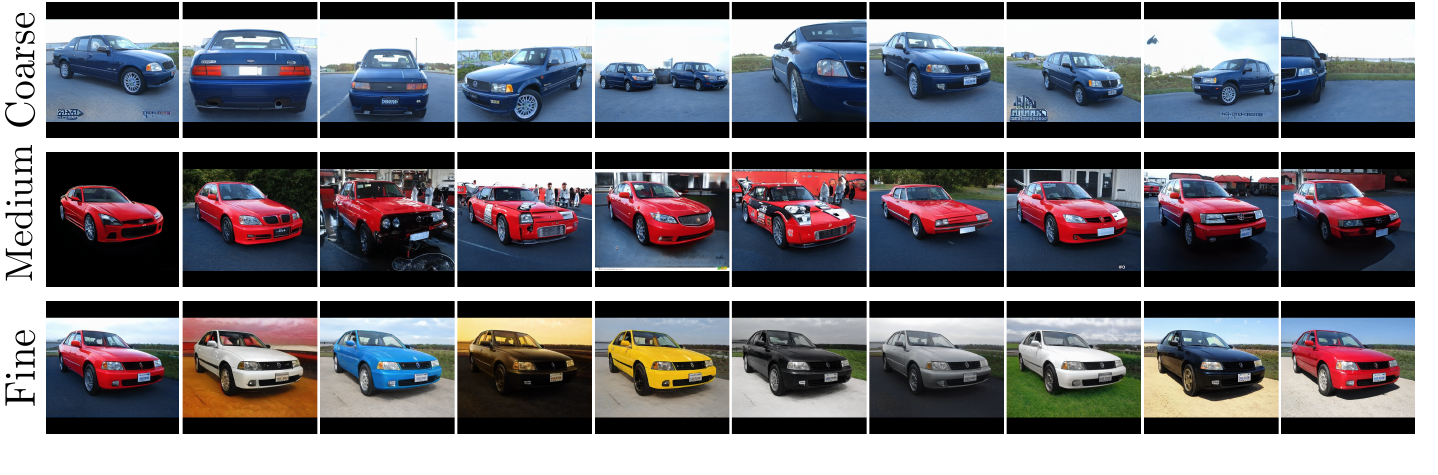}
	\caption{The semantics controlled by different latent levels for StyleGAN trained on the LSUN-cars dataset \cite{kramberger2020lsun}. Coarse layers control the view point and/or car pose, while medium layers control the shape of the car and the background appearance. The fine layers control mainly the colors.}
	\label{fig:high_medium_low_example}
\end{figure*} 
\paragraph{Truncation}
Truncation is a post-training method, sacrificing the variety of a generative model to increase the quality of individual samples. The most well-known truncation method is the ``truncation trick''~\cite{marchesi2017megapixel, brock2018large}: sampling from a truncated normal distribution at test time. This simple procedure had shown to dramatically boost the results in BigGAN~\cite{brock2018large}.
Instead of truncating the distribution, \cite{kingma2018glow} proposed to sample the latent space from a reduced-temperature probability at test time.
Similarly, generative models based on StyleGAN architectures~\cite{karras2019style, karras2020analyzing} adopted a simple truncation operation in $\mathcal{W}$ space. However, instead of truncating the learned manifold in $\mathcal{W}$, Karras \etal~\cite{karras2019style} scale the deviation of specific $w \in \mathcal{W}$ from the global mean by a fixed factor.
Another approach~\cite{kynkaanniemi2019improved} clamps samples in low-density areas to the boundary of high-density areas. Such clamping was shown to outperform the linear interpolation toward the mean.
Assuming that the Frobenius norm of the generator's Jacobian is high in low-density areas, \cite{tanielian2020learning} truncate by rejecting such samples.
The latter is a binary rejection method, whereas we are interested in improving poor samples, rather than discarding them altogether.

\section{Method}
\label{sec:Method}
We propose a new truncation scheme, based on learned clusters in latent space. Given a pre-trained StyleGAN model, we cluster its learned intermediate latent space into $k$ clusters. However, for improved control at different semantic levels, we cluster an extended latent space $\Wspace^L$, where $L \in [1:18]$ (using this notation $\mathcal{W}^1$ is the original $\mathcal{W}$ space and $\mathcal{W}^{18}$ is the $\mathcal{W}^+$ space ~\cite{abdal2019image2stylegan}.
Specifically, we independently cluster each semantic level $l \in [1,\dots, L]$, and preform truncation against the predicted cluster of $w \in \Wspace^l$, which is, by definition, different for every semantic level. Our overall architecture is shown in \figref{fig:arch}. 

\subsection{Semantic clustering of a pre-trained StyleGAN}
\begin{figure}
	\centering
	\includegraphics[width=\linewidth]{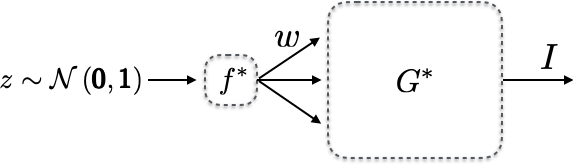}
	\caption{A high-level view of the StyleGAN architecture. A single latent space vector $w \in \mathcal{W}$ generates the style parameters for all the levels of the generator.}
	\label{fig:styleGAN_simple_arch}
\end{figure}

\begin{figure}
	\centering
	\includegraphics[width=\linewidth]{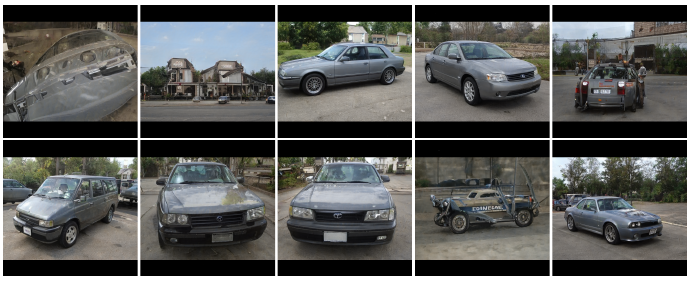}
	\caption{Clustering $\Wspace$ using a Gaussian Mixture Model (GMM). The centers of the clusters reflect that the grouping is according to high-level semantics (view-point and car orientation), rather than according to medium or low-level semantics.}	
	\label{fig:GMM_centers}
\end{figure}

In the StyleGAN architecture, as shown in \figref{fig:styleGAN_simple_arch}, a single $w \in \Wspace$ is fed to all levels of the generator $G$, thus, a single latent vector $w$ controls both high level features and low-level features in the generation process.
Consequently, simple clustering methods in $\Wspace$, such as Gaussian Mixture Models (GMM), typically cluster similar view points or similar scale of the image, but do not reflect any grouping of images based on their medium- or low-level semantic content, as demonstrated in \figref{fig:GMM_centers}.

To learn a more fine-grained semantic clustering, we propose to examine each semantic level $\Wspace^l$ of the extended latent space separately. We model each $\Wspace^l$ as a multi-modal space, and learn a new mapping network $f^l$ to generate it as such. Following \cite{hoang2018mgan}, we feed $f^l$  with $k_l$ learnable gaussians $\left\{\mu^l_i, \sigma^l_i \right\}_{i=1}^{k_l}$, and simultaneously learn to infer the input Gaussian $\pi_i^l$. In practice, $\pi_i^l \in \left\{0,1\right\}$ and we use a classification network $C^l$ to infer which Gaussian generated each $w^l \in \Wspace^l$. Note that the pretrained generator $G$ is kept fixed in this process. Since $G$ was trained to generate images from vectors in $\Wspace$, we use a discriminator $D_w^{l}$, to ensure that $\Wspace^l$ is similar to $\Wspace$.
We adopt WGAN-GP \cite{gulrajani2017improved} as our adversarial loss, and cross-entropy as the classification loss.
The overall loss for layer $l$ is:
\begin{align*}
& \underset{f^l, C^l, \left\{\mu^l_i, \sigma^l_i \right\}_{i=1}^{k_l}}{\min}\underset{D_w^l}{\max}{\quad  E_{w\sim \mathbb{P}_w}\left[D^l_w\left(w\right)\right]}  \\ & - {E_{w^l \sim \mathbb{P}_{f^l}, \pi^l \sim \mathbb{P}_{\pi^l}}\left[ D_w^l\left(w^l\right)-\log\left(\pi^lC^l\left(G\left(w^l\right)\right)\right)\right]} \\ &+ {\lambda_{gp} E_{w\sim \mathbb{P}_w, w^l \sim \mathbb{P}_{f^l} }L_{gp}\left(D, w,w^l\right)},
\label{eq:loss_all}
\end{align*}
where $w^l = f\left(\sum_{i=1}^{k_l}\pi_i^l\mathbb{N}\left(\mu_i, \sigma_i\right)\right)$, $\mathbb{P}_{\pi^l}\left(\pi^l=e_i\right)=1/k_l$ for $i=1,\dots,k_l$ and  $L_{gp}$ is the gradient penalty on straight line between $w$ and $w^l$ as defined in \cite{gulrajani2017improved}.

\subsection{Hierarchy of clusters}
 We train independently for each level $l$ a new mapping function $f^{l}$ with $k_{l}$ clusters, to expose $\sum_l k_{l}$ semantic clusters. While the selection of the levels can be done arbitrarily, since we are interested in semantically meaningful clusters, we have found it beneficial to use only $L=3$ (i.e. we generate in $\Wspace^3$ space), where $l=1$ is the coarse, high-level semantic layer, typically representing view point and scale, $l=2$ is the medium-level semantics, representing background and high shape deformation (e.g. a model of a car) constrained on the coarse layer, and the fine layer, $l=3$, representing low level semantics, such as textures and colors. In Figure~\ref{fig:high_medium_low_example} we demonstrate these three semantic levels for the LSUN-cars dataset. The exact mapping between the StyleGAN generator layers and the three levels of $\Wspace^3$ is reported in the supplementary material.
  
\begin{figure}[t]
	\centering
	\includegraphics[width=\linewidth]{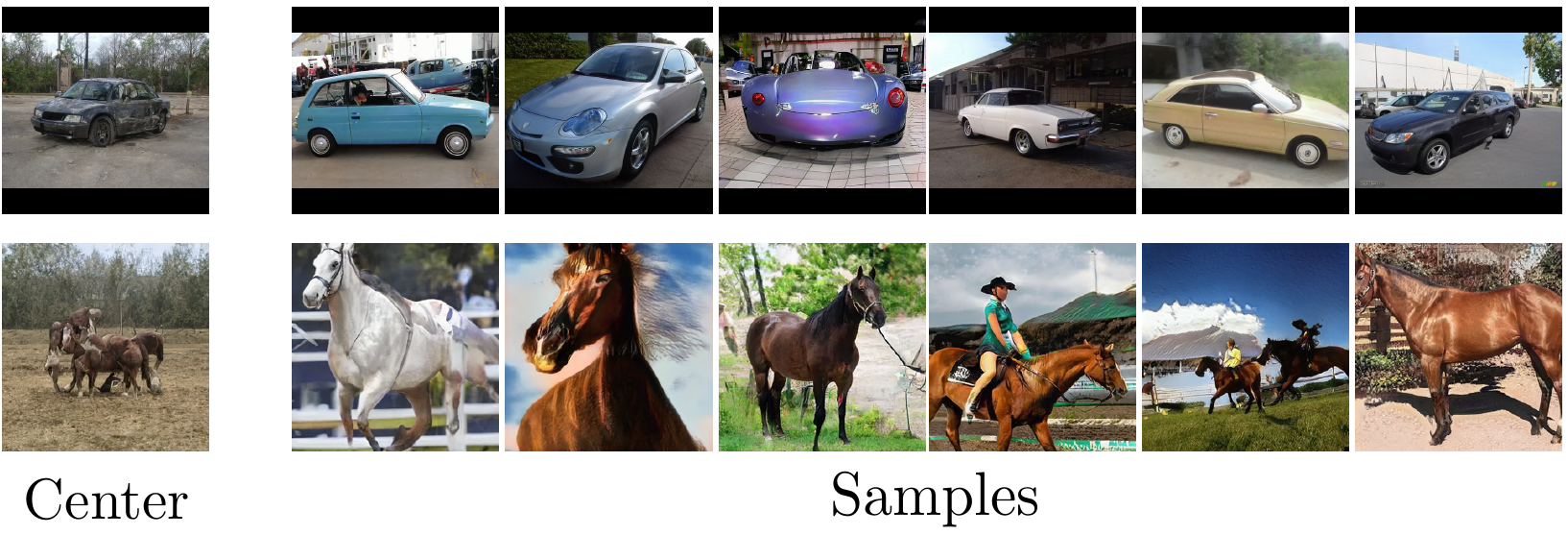}
	\caption{Mean of $\Wspace$ next to several random samples samples for two diverse datasets. When the datasets exhibit great diversity at multiple semantic levels, a single mean can not represent the desired target to interpolate towards, due to significant differences in high-level semantics (view point, shape) as well as low-level semantics (color, texture).}	
	\label{fig:samples_and_centers}
\end{figure}  
\begin{figure}[t]
	\centering
	\includegraphics[width=\linewidth]{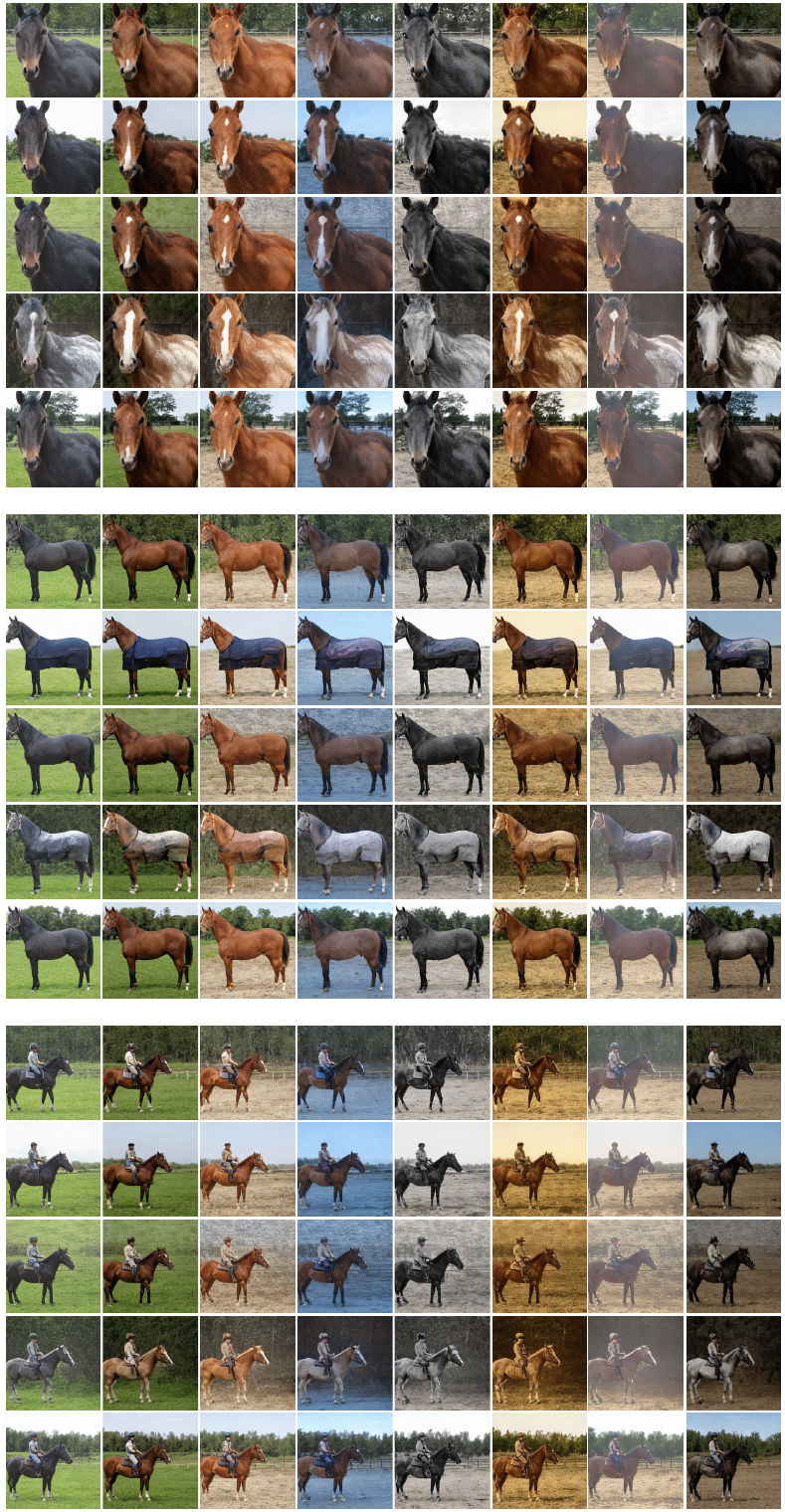}
	\caption{Combinations of different semantic clusters. Each of the three groups of images corresponds to a different high-level cluster. In each group, the rows and columns represent five different medium-level and eight fine-level clusters, respectively. }
	\label{fig:center_horses_prod}
\end{figure} 
 
\subsection{Cluster-based truncation}\label{sec:our-truncation}
The basic truncation scheme used by StyleGAN, assumes that contracting the latent space $\Wspace$ towards the mean improves the output image. This is based on the assumption, that denser areas in the latent space, represent higher quality images. However, while a single mean is perhaps a valid assumption for arguably unimodal datasets such as FFHQ \cite{karras2019style}, when the generated data is multi-modal the assumption does not hold. In Figure~\ref{fig:samples_and_centers} we show the mean and next to it a few random samples for two diverse datasets (LSUN-cars and LSUN-horses), to demonstrate that a single mean cannot represent well the different samples in such datasets.

We propose to utilize the learned clusters in order to develop a novel, more sophisticated, multi-level semantic truncation scheme. 
Given $w \in \Wspace$, we first determine, for each semantic level, which cluster it belongs to. 
Since there are $k_l$ clusters at each level $l$, the total number of possible outcomes (combinations of cluster assignments) is $\prod_l k_{l}$ clusters, as illustrated  in \figref{fig:center_horses_prod}.

Next, for each cluster, we perform a contraction of $w$ towards the corresponding cluster mean, and concatenate them to yield $w_{\mathit{trunc}} \in \Wspace^3$. The truncation procedure is summarized in Alg.~\ref{algo:trunc}.

\begin{algorithm}
\SetAlgoLined
\KwResult{Truncated latent vector $w_{\mathit{trunc}} \in \Wspace^3$ }
\textbf{Input:} Latent vector $w \in \Wspace$, mean of clusters for each level $\left\{\left\{w_{c_i}^l\right\}_{i=1}^{k_l}\right\}_{l=1}^L$, truncation strength $\phi$  \;
\textbf{Initialize:} $w_{\mathit{trunc}} \leftarrow [] $\;

 \For{ l in $[1,\dots,L]$}{
    $i \leftarrow \argmax_i\left\{C^l(w)[i]\right\}$\;
    $w_l \leftarrow w + \phi\left(w^l_{c_i}  - w\right)$ \;
    $w_{\mathit{trunc}} \leftarrow [w_l | w_{\mathit{trunc}}] $\;
 }
 \caption{Multi-level semantic truncation}
 \label{algo:trunc}
\end{algorithm}

\begin{figure*}
	\centering
	\includegraphics[width=\linewidth]{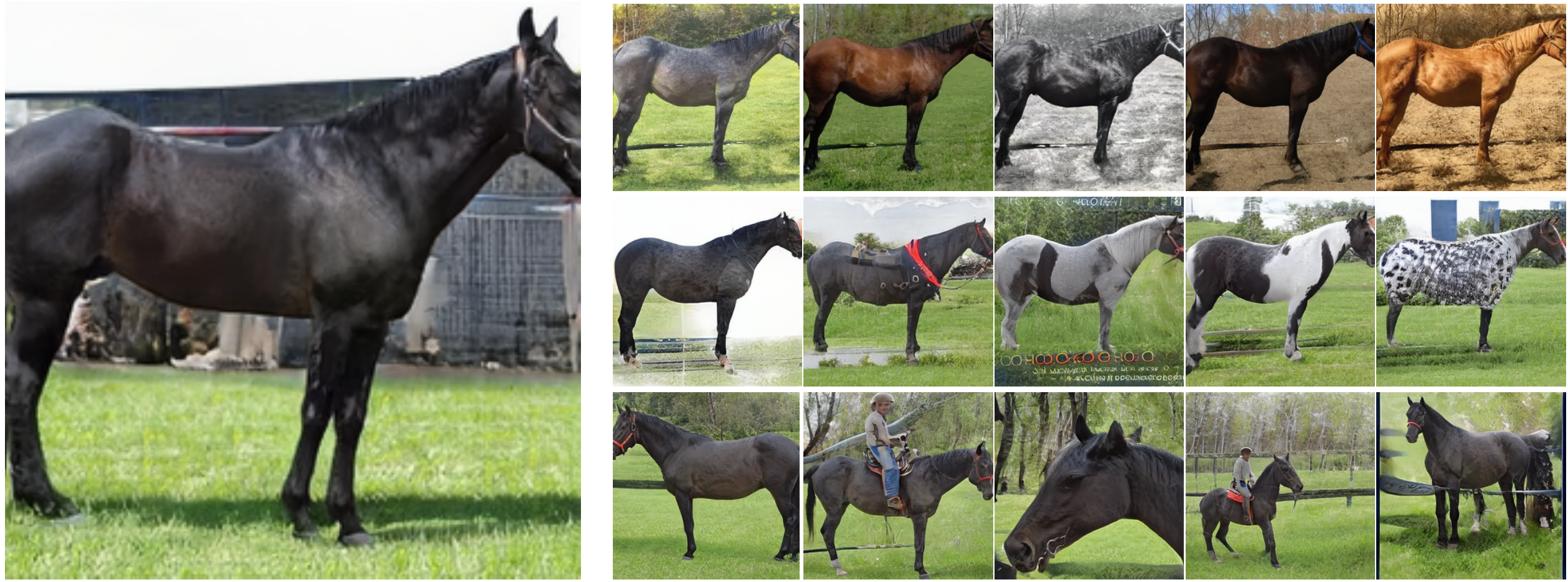}
	\caption{Comparison of styleGAN truncation (upper row) and our semantic truncation (lower row). The images on the left, are truncated using styleGAN center (upper row) and our semantic centers (right). Our truncation generates images, closer to the original image, preserving the viewpoint and the semantic content (number of cars, part of the animal, etc.) }.
\Description[Comparison of styleGAN truncation with our semantic truncation.]{Comparison of styleGAN truncation (upper row) and our semantic truncation (lower row). The images on the left, are truncated using styleGAN center (upper row) and our semantic centers (right). Our truncation generates images, closer to the original image, preserving the viewpoint and the semantic content (number of cars, part of the animal, etc.)}
\label{fig:controlled_generation_2}
\end{figure*}
\section{Experiments}
\label{sec:experiments}

%\begin{figure}
%	\centering
%	\includegraphics[width=\linewidth]{figures/controlled_generation_2.png}
%\caption{Changing clusters in specific levels. Given a generated image (on the left), we are able generate multiple versions of the same image by sampling for a specific levels (From top to bottom: fine, medium and coarse) from different clusters.}
%\label{fig:controlled_generation_2}
%\end{figure} 

\begin{figure}
	\centering
	\includegraphics[width=\linewidth]{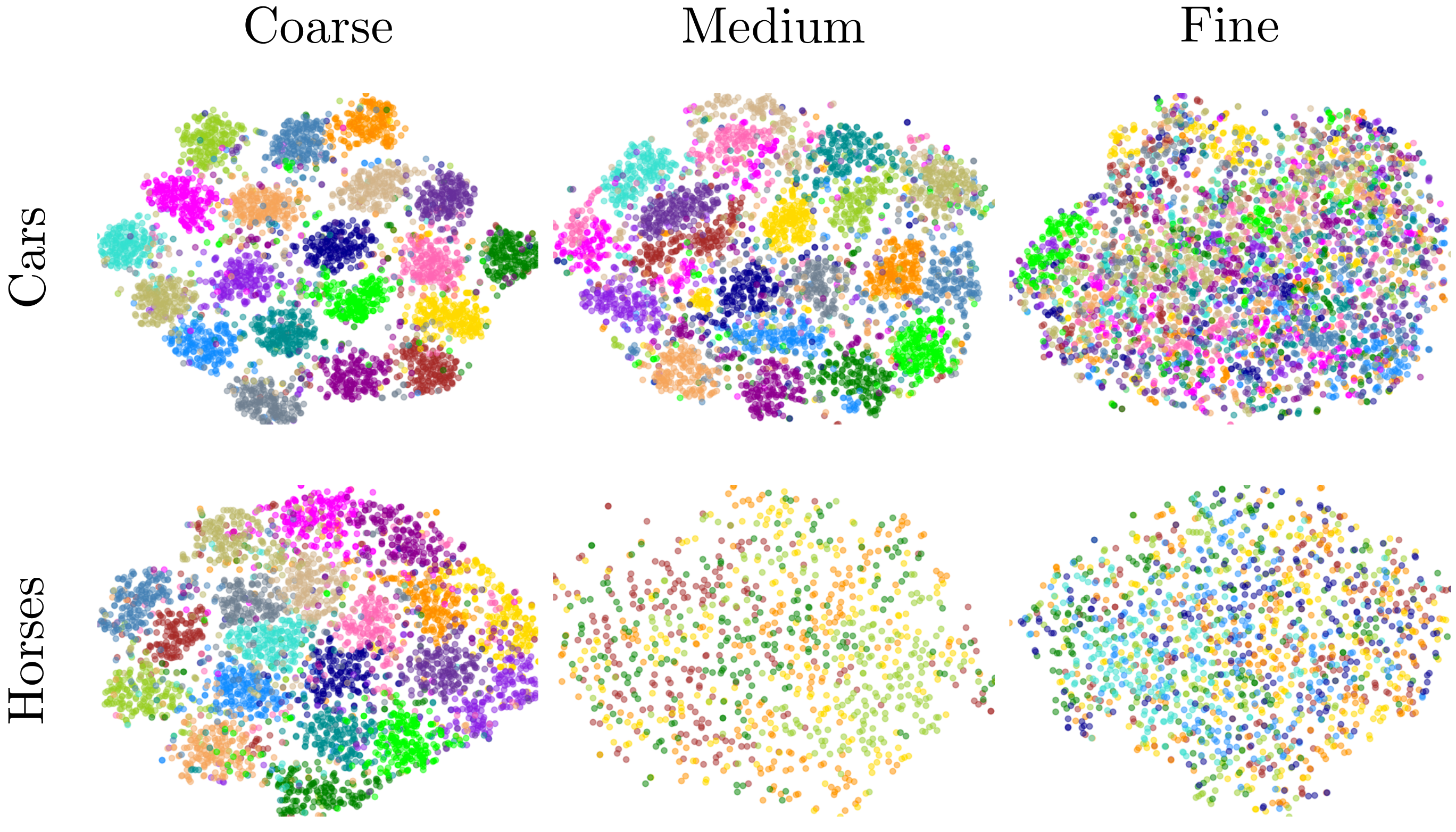}
	\caption{TSNE of $\Wspace$ for different semantic levels. Each color represent different learned cluster by our architecture. While the coarse layer exhibit noticeable clusters, the fine layer does not, indicating that distances in $\Wspace$ has less low level semantic meaning.}.
	\label{fig:TSNE}
\end{figure}

 \begin{figure}[h]
	\centering
	\includegraphics[width=\linewidth]{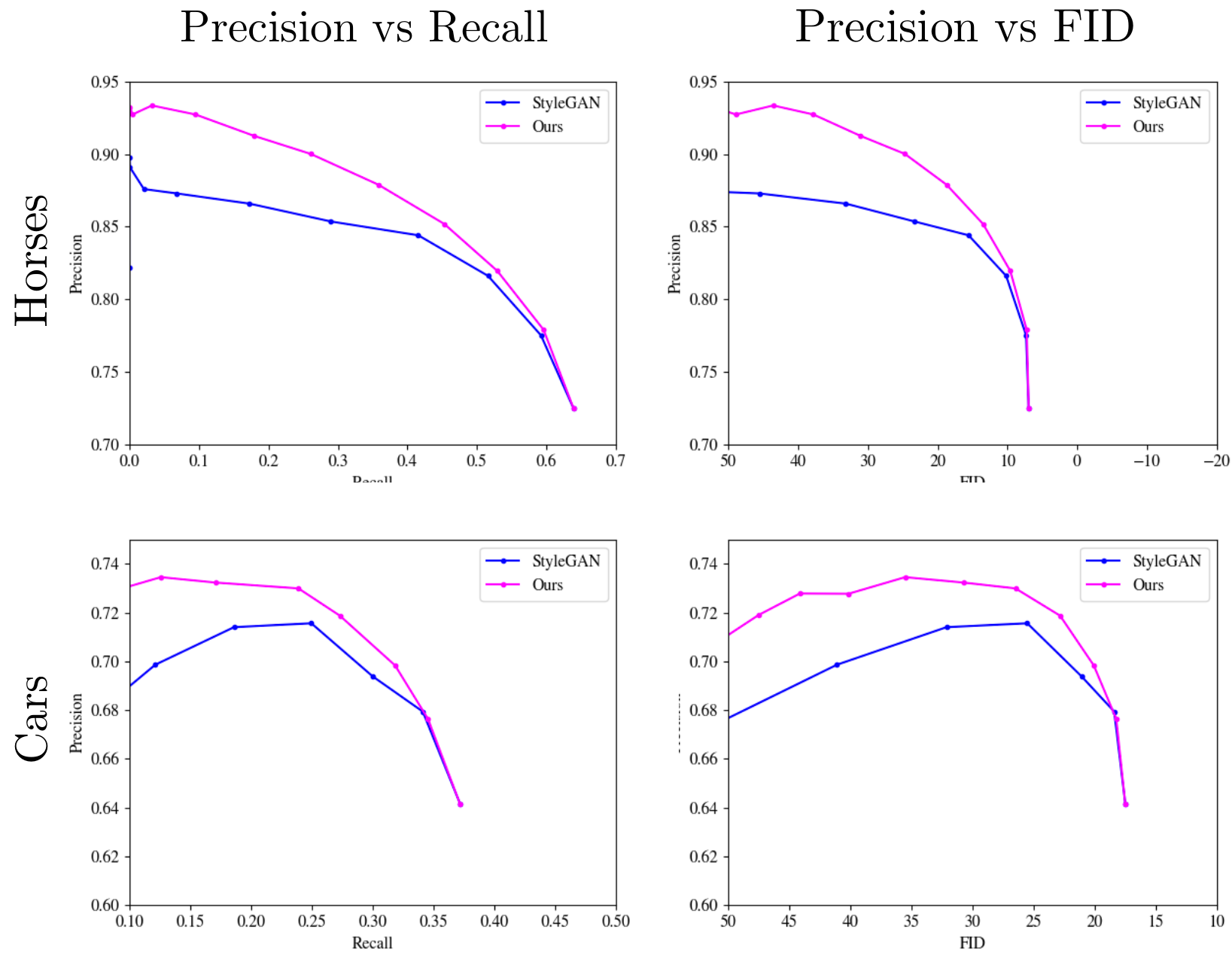}
	\caption{PR curves.}
	
	\label{fig:truncation-comparison-quantity-far}
\end{figure}
\begin{figure*}
	\centering
	\includegraphics[width=\linewidth]{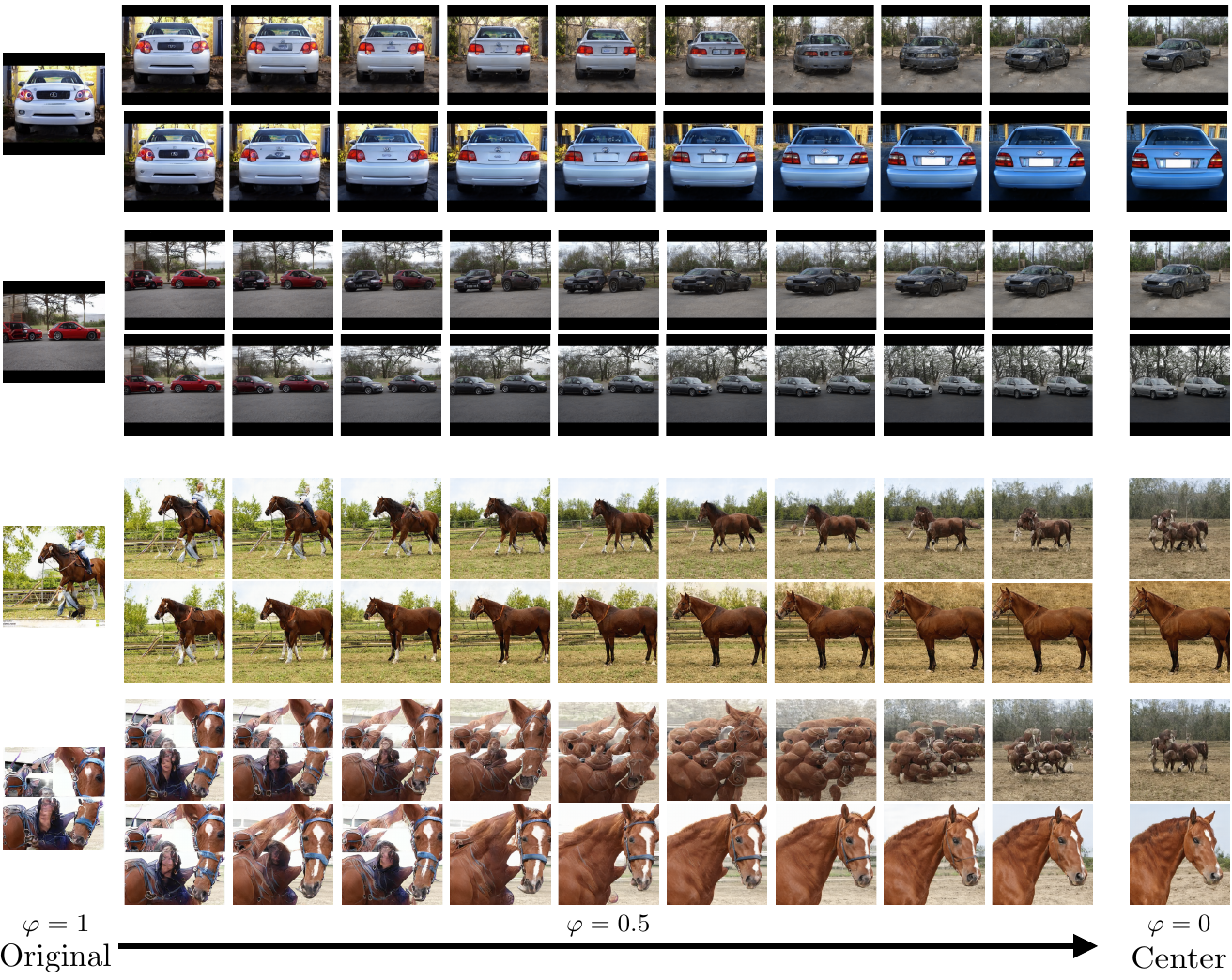}
	\caption{Comparison of styleGAN truncation (upper row) and our semantic truncation (lower row). The images on the left, are truncated using styleGAN center (upper row) and our semantic centers (right). Our truncation generates images, closer to the original image, preserving the viewpoint and the semantic content (number of cars, part of the animal, etc.) }.

\label{fig:truncation-comparison-quality}
\end{figure*}

We evaluate the proposed truncation method on LSUN~\cite{yu2015lsun} horses and cars.  Both datasets exhibit high diversity both in global geometric features (pose, semantic content, background features), and in color and texture (object color, global brightness, texture patterns). In all experiments, we have used the official pre-trained StyleGAN2 generator~\cite{karras2020analyzing}. 
Since these pre-trained generators, were trained to generate different sizes of images ($256 \times 256$ for horses and $512 \times 512$ for cars) , they have a different amount of convolution layers ($14$ and $16$ respectively). 
In all our experiments, we have used $3$ semantic levels $w^l \quad (l=1\dots 3)$.
Our coarse mapping function, generating $w^0$ feeds the first $4$ convolution layers of the StyleGAN2 generator. The medium semantic level mapping function is fed to the next $4$ convolution levels. The fine mapping function is fed to the remaining layers, but the last one. We have found that assigning the last layer of StyleGAN2 generator to our fine level $w^2$ leads to artifacts, thus, we feed the last convolution layer with vector $w \in \Wspace$ from the original styleGAN mapping function, thereby limiting our fine mapping function to control only $7$ and $5$ layers for the cars and horses generators, respectively.
We have trained our multi-modal mapping functions, classifier, and discriminator for $10000$ iterations, with batch size $4$, where all semantic levels for a specific dataset were trained independently but simultaneously. 

\subsection{Multi-level controlled generation}
Our learned mapping functions per level, also enable us to control the generation process. Specifically, we can alternate at a specific level between different clusters. As these clusters are from the same level, the difference between them is limited to specific attributes.  In \figref{fig:controlled_generation_2} we show an example of such control. We first generate a specific $w^0, w^1, w^2$ for the coarse, medium, and fine levels respectively, from a specific cluster in each level, yielding a single image. Then, we sample a new latent code from a different cluster in a single level, while fixing other levels' latent codes. The result is a different version of the same image, corresponding to the control of the levels and the learned clusters \figref{fig:controlled_generation_2}.

%\begin{figure}[t]
%	\centering
%	\includegraphics[width=\linewidth]{figures/centers_coarse_horse.png}
%	\caption{centers horses.
%	}
%	\label{fig:centers_coarse_horse}
%\end{figure}

\subsection{Evaluating the modalities}
Our learned modalities are driven by a classifier in the image domain, while our architecture drives each level of clusters to capture modalities at different semantic level. In \figref{fig:TSNE} we observe the correlation between these clusters and the Euclidean distance in $\Wspace$. We randomly select $1000$ latent vector $z$ of every cluster and feed forward them through our learned mapping function. We then  reduce the dimension by t-SNE, separately for each level. As can be seen in \figref{fig:TSNE}, the coarse level exhibit high correlation in $\Wspace$ (i.e. clusters can be observed), while the fine level clusters are not observable in $\Wspace$ at all. The meaning of this results, is that $\Wspace$ is not a good space to measure the distance between low level semantic.

\subsection{Truncation}
We leverage our learned latent clusters to preform our multi-level  truncation procedure, as described in Section \ref{sec:our-truncation} and \algoref{algo:trunc}.
We calculate, for each cluster $i$ at level $l$ it's center $i$, $w_{c_i}^l$ by clustering $N=10000$ samples of $w \in \Wspace$. To evaluate the quality of our truncation methods, we preform truncation at different "strength" and evaluate the quality of the generated samples.

\paragraph{Quantitative comparison.} 
To quantitatively compare our method with StyleGAN we follow the procedure proposed in \cite{kynkaanniemi2019improved}. We first randomly sample $10,000$ images from the dataset. Similarly, we generate the same amount of latent vectors $w \in \Wspace$,
and their corresponding images, $I=G(w)$. We then cluster each $w$ at every level (i.e. coarse, medium and fine).
The "strength" of the truncation is varied by sampling $11$ values in the range of $\varphi \in \left(0,1\right)$. For each value we truncate the pre-sampled $w \in \Wspace$, using both StyleGAN truncation and our semantic truncation and generate the corresponding images. These generated images are then feed-forwarded through Inception-v3 \cite{szegedy2016rethinking} pre-trained on ImageNet until the last fully-connected layer.
For each $\varphi$ we calculate the precision, recall and FID \cite{heusel2017gans} as defined in \cite{kynkaanniemi2019improved}, and accumulate the results to a PR-curve and a P-FID curve as shown in Figure~\ref{fig:truncation-comparison-quantity-far}. As can be seen, our truncation achieves higher precision for the same FID and Recall compared with the standard StyleGAN truncation.

\paragraph{Qualitative comparison.} 
We show the quality of our truncation method, compared to the original StyleGAN truncation in \figref{fig:truncation-comparison-quality}. While StyleGAN truncation does improve the quality of the results, the contraction to a specific mean, reduces the variability, both for high-level semantics (shape, scale) and low-level semantics (background, color). Our truncation on the other hand allows a more detailed approach, where first the appropriate center is selected for each level, thus enabling more variability in each level, but moreover, most of the truncation results, while improving the quality, remain faithful to the original image in term of the pose, semantic content, background, and color as shown in Figure~\ref{fig:truncation-comparison-quality}.

%We compare in \figref{fig:truncation-comparison-quantity} the precision, recall and FID of all truncated versions  as defined in  of our truncation with StyleGAN truncation \cite{karras2019style}. As can be seen, our method is better in both precision, recall, and FID. Perhaps a more revealing experiment is to examine the effect of truncation only on low probability latent vectors in $\Wspace$. To do so, we sample a large number of latent vectors $w \in \Wspace$ and filter only the $N=10000$ farthest from the center. In \figref{fig:truncation-comparison-quantity-far} we show the comparison conducted only for these less likely latent vectors. Here the advantage of  our method over the naive StyleGAN truncation is more evident, as the precision increases rapidly harming the recall only slightly. 

\section{Conclusions}
\label{sec:conclusion}
We proposed a new truncation technique, based on a decomposition of the latent space into clusters, at different levels of the extended latent space $\Wspace^L$. We showed that such truncation improves previous results both qualitatively and quantitatively, producing faithful output to the original un-truncated image. Our re-arrangement of the latent space also allows a more controlled generation by selecting the desired cluster at each level. We believe that this new arrangement paves future research for methods that aim to edit images by inverting to styleGAN latent space. As editing requires an inversion mechanism to generate a single $w \in \Wspace$ for all levels, our method may remove such constraint by requiring only inversion to different clusters at different levels.

\bibliographystyle{ACM-Reference-Format}
\bibliography{ref}
\clearpage
\end{document}